\documentclass[10pt,twocolumn,letterpaper]{article}

\usepackage{cvpr}              

\usepackage{graphicx}
\usepackage{amsmath}
\usepackage{amssymb}
\usepackage{booktabs}
\usepackage{gensymb}
\usepackage{comment}
\usepackage{braket}
\usepackage{bm}

\usepackage[pagebackref,breaklinks,colorlinks]{hyperref}

\usepackage[capitalize]{cleveref}
\crefname{section}{Sec.}{Secs.}
\Crefname{section}{Section}{Sections}
\Crefname{table}{Table}{Tables}
\crefname{table}{Tab.}{Tabs.}

\def\cvprPaperID{37} 
\def\confName{CVWW}
\def\confYear{2024}

\newcommand\norm[1]{\lVert#1\rVert}

\begin{document}

\date{}
\title{Video shutter angle estimation using optical flow and linear blur}

\author{David Korčák\qquad\qquad Jiří Matas\\
Faculty of Electrical Engineering\\
Czech Technical University in Prague\\
{\tt\small korcadav@fel.cvut.cz  \qquad matas@fel.cvut.cz} \\
}
\maketitle

\begin{abstract}
We present a method for estimating the shutter angle, a.k.a. exposure fraction -- the ratio of the exposure time and the reciprocal of frame rate -- of videoclips containing motion. The approach exploits the relation of the exposure fraction, optical flow, and linear motion blur. Robustness is achieved by selecting image patches where both the optical flow and blur estimates are reliable, checking their consistency. The method was evaluated on the publicly available Beam-Splitter Dataset with a range of exposure fractions from 0.015 to 0.36. The best achieved mean absolute error of estimates was 0.039.
We successfully test the suitability of the method for a forensic application of detection of video tampering by frame removal or insertion.

\end{abstract}

\section{Introduction}
\label{sec:intro}

The shutter angle, a.k.a the exposure fraction, is the ratio of the exposure time, i.e. the time period a film or a sensor is exposed to light, and the time between two consecutive frames, i.e. the reciprocal of the frame rate.
The shutter angle determines the relation between object motion and  image blur and thus influences viewer perception.
This has been used in film-making as part of artistic expression  and tutorials and websites are devoted to explaining the effect \cite{red_camera_shutter}. In computer vision and image processing, exposure fraction affects methods for temporal super-resolution and video frame interpolation, since the inserted images need to both interpolate motions and reduce motion blur. 

For analog cameras, the exposure fraction remains constant throughout the duration of the video. In digital cameras, the exposure time and thus the shutter angle may be set dynamically, according to illumination intensity. Nevertheless, for most recorded videos it stays constant. For global shutter cameras, it is the same for every pixel of a frame. For rolling shutter cameras the same is true for horizontal motions; the exact modeling is more complex for vertical motions.

The exposure fraction  provides a physics-based constraint that influences every pixel, and it thus has the potential for the detection of fake videos and local image edits.  Violations of the constraint -- the linear relationship between the local motion blur and optical flow with the shutter angle providing the scaling constant -- are not immediately obvious to human observers, and thus might not be noticed by neither the authors nor the viewers of the altered or synthesized content. Moreover, many generators of synthetic content are often trained on sharp images corresponding to very short exposures or on graphics-generated data and thus might not represent motion blur as required by physics.

In the paper, we present a robust  method for the estimation of the shutter angle that relies on explicitly running a state-of-the-art optical flow algorithm \cite{DBLP:journals/corr/abs-2003-12039} and linear blur kernel estimator \cite{gong2017blur2mf}.  We are not aware of any existing method for shutter angle estimation from unconstrained video sequences. Barber \etal \cite{BARBER2015155}
 formulate the problem for sequences containing only specific types of blur. We are the first to address the problem for sequences containing general motion.

 
 In summary, we make the following contributions. The proposed method is novel, exploiting recent progress in deep optical flow and linear blur estimation. Both of these estimates are dense, permitting to achieve robustness by combining predictions from patches where both estimates are reliable and consistent.
We  show a forensic application of the method, considering detection of video tampering by frame removal or insertion.
\section{Related work}
\label{sec:related_work}

The problem of estimating the shutter angle of a videoclip has been approached from the point of view of precisely measuring camera characteristics with the help of a bespoke setup. 
The method of Simon et al. \cite{sensors} relies on  special devices, such as turntables, CRT displays or arrays of LEDs lit in specific patterns. Barber \etal \cite{BARBER2015155} addressed the problem for sequences containing blur induced by either zoom or camera rotation during exposure.

We formulate the problem of estimating video shutter angle with the use of optical flow and linear blur kernel estimates from a general video clip containing motion.
Methods for estimating linear blur parameters \cite{gong2017blur2mf, exp_traj, sunetal} often rely on deep neural networks trained on synthetically blurred images. Typically, they  serve as an intermediate step towards image or video frame deblurring. While the details of particular implementations differ, the output is generally a set of estimated parameters of linear blur kernels.

Similarly, the topic of optical flow estimation has seen a lot of progress with methods such as Teed and Deng \cite{DBLP:journals/corr/abs-2003-12039}'s RAFT. Optical flow methods seek to estimate pixel displacements between consecutive frames, and as of recently, are based on various deep neural network architectures. As obtaining ground truth of optical flow on real-world data is difficult, these methods are trained on synthetic datasets.

\section{Method} 
\label{sec:shutter_estimation}
The proposed method for shutter angle estimation relies on the calculation of dense optical flow, described in  \cref{subsec:optical_flow}, and linear blur, described in \cref{subsec:linear_blur}. The shutter angle is the ratio of the length of blur and optical flow vectors (\cref{subsec:estimate_frac}) if the direction of motion does not change within the exposure time, thus not all parts of the image are suitable for estimating this ratio. For instance, both estimates of the blur and the optical flow will be unreliable in parts of the image that contain the sky or similar smooth-texture surfaces. In \cref{subsec:computation} we describe our algorithm for the selection of patches suitable for shutter angle estimation. 

\subsection{Optical flow model}
\label{subsec:optical_flow}
Optical flow is conventionally defined as per-pixel motion between video frames. Given 2 consecutive video frames $F_{i}$ and $F_{i+1}$, the goal of optical flow estimation methods is to map the position $(x_{i},\,y_{i})$ of a given pixel in frame $F_{i}$ to its position $(x_{i+1},\,y_{i+1})$ in frame $F_{i+1}$. This mapping can be modeled as a dense displacement field $(\mathbf{f^{1}},\,\mathbf{f^{2}})$. The position of a given pixel in frame $F_{i+1}$ can be then described as $(x_{i} + f^{1}(x_{i}),\,y_{i}+f^{2}(y_{i}))$\cite{DBLP:journals/corr/abs-2003-12039}. In this paper, we employ Teed and Deng's RAFT \cite{DBLP:journals/corr/abs-2003-12039} for optical flow estimation, selected for its well-documented performance across various sequences \cite{DBLP:journals/corr/abs-2003-12039} and strong benchmark results \cite{ZHAI2021107861}.

\subsection{Linear blur model}
\label{subsec:linear_blur}

The heterogeneous motion blur model commonly views the blurred (real) image $\mathbf{Y}$ as the product of a convolution of a sharp image $\mathbf{X}$ with an operator $\mathbf{K}$ and additive noise $\mathbf{N}$ \cite{gong2017blur2mf}.
\begin{equation}
    \mathbf{Y} = \mathbf{K}\,*\,\mathbf{X} + \mathbf{N}
\end{equation}
The motion blur kernel map, $\mathbf{K}$, in general, consists of different blur kernels for each pixel at position $(x, \,y)$.  

The linear blur assumes the kernels at $(x, \,y)$ can be modeled as 1D, in the direction of the local motion. 
Such kernels can also be interpreted as two-dimensional motion vectors $\mathbf{K_{x,\,y}} = (\mathbf{k^1_{x,\,y}},\mathbf{k^2_{x,\,y}})$. 
Linear blur kernels also characterize the motion of a pixel over the camera exposure time $\varepsilon$, as the blurring occurs by motion during camera light capture over the exposure period.

The assumption of linear blur is violated e.g. for hand-held cameras that may undergo Brownian-like motions.  
In such cases, the estimation of both blur kernels and optical flow is difficult and they present a challenge for our method. 
Since the exposure fraction is the same for all pixels in the image, it is sufficient to find a modest number of areas where the linear blur assumption holds, e.g. on a linearly moving object in the scene. In \cref{subsec:computation}, we introduce techniques for identifying and selecting such areas of video frames.

In this paper, we apply the method of linear blur estimation by Gong \etal \cite{gong2017blur2mf}, which shows both good generalization ability and dataset performance. It is also one of the only methods that perform per-pixel estimates of blur kernels, i.e. the estimates are calculated, not interpolated from patches, in full resolution. This method, however, introduces a level of discretization error, as the deep neural network used for estimating blur kernels operates as a multi-class classifier with a discrete output space. We attempt to minimize the effect of such errors on our final estimate as described in \cref{subsec:computation}.

During testing, we also evaluated the performance of Zhang \etal's method \cite{exp_traj} as it operates with a real output space but found that in our set up it performed worse than Gong \etal's method \cite{gong2017blur2mf}.

\subsection{Shutter angle from linear blur and optical flow}
\label{subsec:shutter_formulation}
We consider a video camera with the following parameters. Let $\varepsilon$ denote the exposure time of each frame, $f$ the video framerate in frames per second and $\theta$ the video shutter angle in degrees. For the i-th video frame, we define $t_{i}$ as the time of exposure start and $t_{i}+\varepsilon$ as the exposure end. 
The time difference between exposure starts of two consecutive frames,
$\Delta t = t_{i+1}\, -\,t_{i} = 1/f = f^{-1}$,
 is equal to the reciprocal of the frame rate $f$.

For the purpose of simplicity, we use the exposure fraction notation, rather than the shutter angle, i.e. instead of  180$\degree$ we speak of 0.5. The degree notation is widely used in cinematography, as it originates from the construction of historical cameras that utilized mechanical rotating shutters to set the exposure time $\varepsilon$. In many modern digital cameras, exposure time $\varepsilon$ can be set explicitly. We  define \textbf{\textit{exposure fraction} $\alpha$} as the ratio 
\begin{equation}
        \alpha = \dfrac{\varepsilon}{\Delta t} = \dfrac{\theta}{360 \degree }.
        \label{eq:alpha}
\end{equation}

 \subsection{Estimating  the exposure fraction}
 \label{subsec:estimate_frac}
Consider optical flow described in \cref{subsec:optical_flow} and linear blur kernel described in \cref{subsec:linear_blur}. The magnitude of the optical flow vector $\norm{\mathbf{f_{x,\,y}}}$ is equivalent to the distance displaced by pixel over the duration of a single frame, i. e. over the time interval of length $\Delta t$. Similarly, the magnitude of linear blur kernel $\norm{\mathbf{K_{x,\,y}}}$ corresponds to the distance displaced by pixel $(x, \,y)$ over the time interval $\varepsilon$. Here, we assume uniform motion, i. e. the pixel at $(x, \,y)$ is traveling at a constant velocity $v_{x,\,y}$ between consecutive frames. Such assumption is reasonable, as the absolute frame duration $\Delta t$ of video clips shot at multiple frames per second is often negligible compared to camera motion or motion of common objects. Under this assumption, we may express the norm of optical flow vector and linear blur kernel as distance displaced by pixel $(x, \,y)$ at constant velocity $v_{x,\,y}$ over respective time intervals.
\begin{equation}
    \norm{\mathbf{f_{x,\,y}}}_{2} = v_{x,\,y}\cdot \Delta t, \norm{\mathbf{K_{x,\,y}}}_{2} = v_{x,\,y}\cdot \varepsilon
    \label{eq:blur_flow_velocity}
\end{equation}
We substitute in \cref{eq:alpha} and obtain the following
\begin{equation}
    \alpha_{x,\,y} = \dfrac{\norm{\mathbf{K_{x,\,y}}}_{2}}{\norm{\mathbf{f_{x,\,y}}}_{2}}
    \,\,;
    \label{eq:alpha_estimation}
\end{equation}

i.e. given the magnitude of the optical flow $\mathbf{f_{x,\,y}}$ and linear blur kernel $\mathbf{K_{x,\,y}}$ at pixel $(x, \,y)$ 
the value of $\alpha$ at position $(x, \,y)$ as a ratio of magnitudes of the two vectors.

\subsection{Computation}
\label{subsec:computation}
The proposed method for estimating the value of $\alpha$ builds on \cref{subsec:estimate_frac}.
As described in \cref{sec:intro}, modern video cameras operate either in a global shutter mode, where all pixels of the frame get exposed at the same point in time or, more commonly,  in a rolling shutter mode, where exposure is performed row-wise. In both cases, the time of exposure $\varepsilon$ remains constant for all pixels in the frame. Similarly, the time interval between exposures of pixels (both global and rolling shutter) remains constant. As a result of these physical constraints, the value of $\alpha$ must be consistent in an entire frame, and typically in the entire video clip. Therefore, the problem of estimating the value $\alpha_{x,\,y}$ pixel-wise is reduced to estimating one global value $\alpha_{\text{glob}}$ for the entire video clip.

As the sources of both optical flow and linear motion blur are not robust and prone to errors in their estimates, and the condition of motion in a single direction during exposure time may not be satisfied, the proposed method locates patches of pixels with the lowest error potential in both linear blur kernels and optical flow. We define multiple constraints and show that are effective in filtering erroneous predictions.

First, we discard estimates at all positions $(x, \,y)$, where the angle between the linear blur kernel and optical flow vectors exceeds a threshold. The condition is evaluated in the cosine domain which avoids wrap-around effects and also addresses the problem that the blur kernel is estimated modulo $\pi$, it does not have a direction:
\begin{equation}
\label{eq:cosine_constraint}
      \dfrac{|\braket{\mathbf{f_{x,\,y}|\mathbf{K_{x,\,y}}}}|}{\norm{\mathbf{f_{x,\,y}}}_2\cdot\norm{\mathbf{K_{x,\,y}}}_2} \geq \cos{\varphi}
\end{equation}
where $\varphi$ is the maximum angle threshold in degrees; $\braket{\cdot|\cdot}$ denotes the dot product.

Second, for \cref{eq:alpha_estimation}   it follows that the norm of the ground truth optical flow vector is always larger than the norm of the ground truth linear blur kernel -- a pixel cannot be physically captured for a longer period than the maximum inter-frame period $\Delta t$. We therefore remove all values from positions  $(x, \,y)$ where:
\begin{equation}
    \norm{\mathbf{K_{x,\,y}}}_2 >  \norm{\mathbf{f_{x,\,y}}}_2
    \label{eq:blur_flow_length_constraints}
\end{equation}

The blur kernel estimation method \cite{gong2017blur2mf}  outputs values in a discrete domain, with the discretization error introduced in both vertical and horizontal directions equal to 1. This renders predictions with small motions arbitrarily, and we thus remove positions with small flow and blur magnitudes:
\begin{equation}
    \norm{\mathbf{K_{x,\,y}}}_2 \leq  1,\, \norm{\mathbf{f_{x,\,y}}}_2 \leq 1  \,\, .
    \label{eq:minimum_length_constraint}
\end{equation}

Next, we find a $D$ x $D$ patch that contains the highest number of valid positions.  The value of $\alpha$  for a given frame is estimated from this patch. 
We calculate the estimate of $\alpha = \alpha_{\text{patch}}$ for the current frame as the ratio of norms of means of linear blur kernels and optical flow vectors
\begin{equation}
    \alpha_{\text{patch}} = \dfrac{\norm{\frac{1}{N} \sum\limits_{i=1}^{N}\mathbf{K_{x_{i},\,y_{i}}}}_2}{\norm{\frac{1}{N} \sum\limits_{i=1}^{N}\mathbf{f_{x_{i},\,y_{i}}}}_2}
\end{equation}
where $N$ is the number of valid positions $(x, \,y)$ inside of the selected patch.

Finally, we calculate $\alpha_{\text{glob}}$ as the median of estimates of all individual frames
\begin{equation}
    \alpha_{\text{glob}} = \text{med}\{\alpha_{\text{patch}_{1}},\,\alpha_{\text{patch}_{2}},\,...,\,\alpha_{\text{patch}_{N}} \}
\end{equation}
where $N$ is the number of frames in the video clip.

\begin{table}
\renewcommand{\arraystretch}{1.2}
  \centering
  \begin{tabular}{ccccccc}
    \toprule
    $\varepsilon$ ($ms$) & 1 & 2 & 3 & 8 & 16 & 24  \\
    \midrule
    $\alpha$ & $0.015$ & $0.030$ & $0.045$ & $0.120$ & $0.240$ & $0.360$\\
    \bottomrule
  \end{tabular}
  \caption{Exposure fractions of BSD subsets based on exposure time $\varepsilon$. All videoclips have a framerate $f = 15$ FPS, $\Delta t = 0.0\overline{66}\,s$ }
  \label{tab:bsd1}
\end{table}

\begin{table*}
\renewcommand{\arraystretch}{1.2}
\centering
  \begin{tabular}
  {ccccccccc}
    \toprule
     &$\varepsilon$ (ms)  & 1 & 2 & 3 & 8 & 16 & 24 & Average\\
    \midrule
     $D$ &$\varphi$   &  &  &  &  &   &  & \\
    \midrule
    $10$ & $3\degree$ & 0.058 & 0.034 & 0.035 & 0.024 & 0.048 & 0.091 & 0.049 \\
    $10$ & $5\degree$ & 0.054 &  0.030 & 0.033 & 0.025 & 0.052 & 0.095 & 0.048\\
    $10$ & $7\degree$ & 0.051 & 0.030 & 0.032 & 0.026 & 0.055 & 0.096 & 0.048\\
    \midrule
    $20$ & $3\degree$ & 0.054 & 0.033 & 0.031 & 0.023 & 0.041 & 0.080 & 0.044\\
    $20$ & $5\degree$ & 0.048 & 0.029 & 0.029 & 0.022 & 0.042 & 0.081 & 0.042\\
    $20$ & $7\degree$ & 0.046 &  0.029  & 0.029 & 0.026 &0.043 & 0.082 & 0.042\\
    \midrule
    $30$ & $3\degree$ & 0.054 & 0.033 & 0.031 & 0.022 & 0.038 & 0.077 & 0.042\\
    $30$ & $5\degree$ & 0.047 & \textbf{0.028} & \textbf{0.028} & \textbf{0.020} & 0.035 & 0.075 & \textbf{0.039}\\
    $30$ & $7\degree$ & \textbf{0.045} & 0.029 & 0.029 & 0.025 & \textbf{0.034} &  \textbf{0.072} & \textbf{0.039}\\
    \bottomrule
  \end{tabular}
  \caption{BSD dataset - mean absolute error of exposure fraction $\hat{\alpha}$ estimates for a range of patch sizes $D$ and the tolerated angular difference~$\varphi$ between the optical flow and blur directions. The best results, in bold, were achieved for the largest window size and angular tolerance of 5-7 $\degree$.  } 
  \label{tab:bsd_results_all}
\end{table*}

\section{Experiments and evaluation}
\label{sec:eval}

In this section, after presenting the values of the two parameters of the proposed method - the patch size $D$ and the angular threshold $\varphi$, we perform testing on a public dataset and in-depth experiments on individual video clips. We investigate both well-performing video clips and failure cases in an attempt to find the limitations of the proposed method. 

\subsection{Parameter selection}
\label{subsec:params}
We tested all configurations with patch sizes of $D = \{10,\,20,\,30\}$ and angular constraints, $\varphi = \{3\degree$,\, $5\degree$,\,$7\degree\}$. For optical flow estimates, we utilized RAFT model \cite{DBLP:journals/corr/abs-2003-12039} pretrained on the Sintel dataset with 12 iterations per two consecutive frames. The results are summarized in \cref{tab:bsd_results_all}.

\begin{figure*}
  \centering
  \begin{subfigure}{0.33\linewidth}
    \includegraphics[width=6cm]{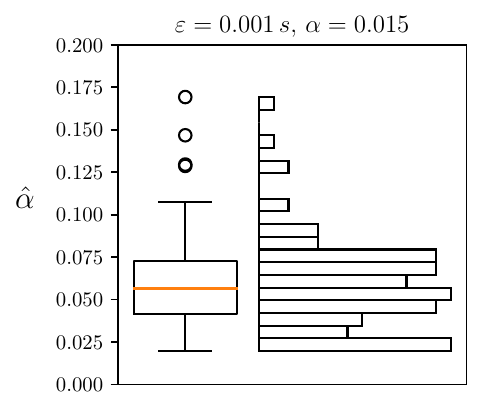}
    \label{fig:short-a}
  \end{subfigure}
  \hfill
  \begin{subfigure}{0.33\linewidth}
    \includegraphics[width=6cm]{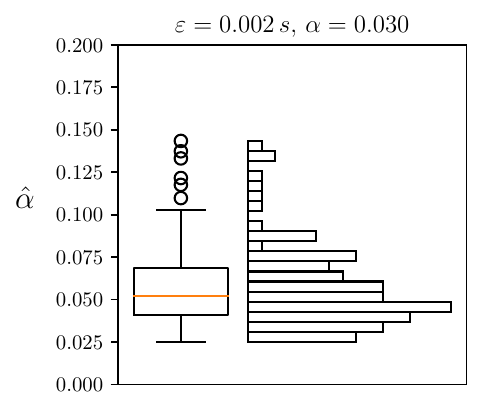}
    \label{fig:short-b}
  \end{subfigure}
  \hfill
     \begin{subfigure}{0.33\linewidth}
    \includegraphics[width=6cm]{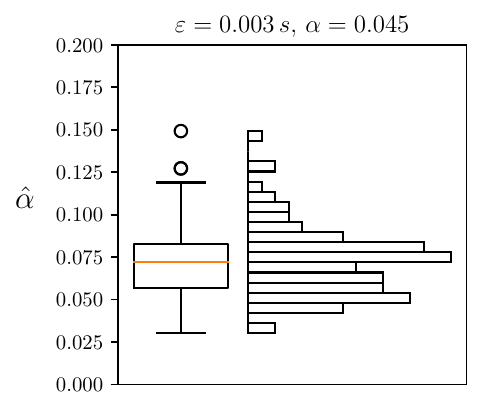}
    \label{fig:short-b}
\end{subfigure}
 \caption{Histograms and box plots of $\hat{\alpha}$ estimates on clips from the BSD dataset with $\varepsilon = \{0.001\,s ,\,0.002\,s,\,0.003\,s\}$. Estimation parameters $\varphi = 5\degree$, $D = 30$. Note that only the (0, 0.2) range is displayed.}
  \label{fig:short_estim}
\end{figure*}

\begin{figure*}
  \centering
  \begin{subfigure}{0.33\linewidth}
  \centering
    \includegraphics[width=6cm]{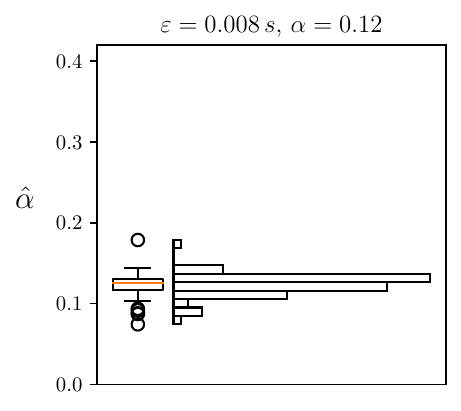}
    \label{fig:long-a}
  \end{subfigure}
  \hfill
  \begin{subfigure}{0.33\linewidth}
    \includegraphics[width=6cm]{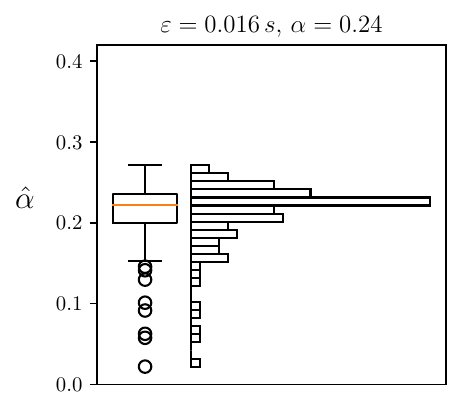}
    \label{fig:long-b}
  \end{subfigure}
  \hfill
  \begin{subfigure}{0.33\linewidth}
  \centering
    \includegraphics[width=6cm]{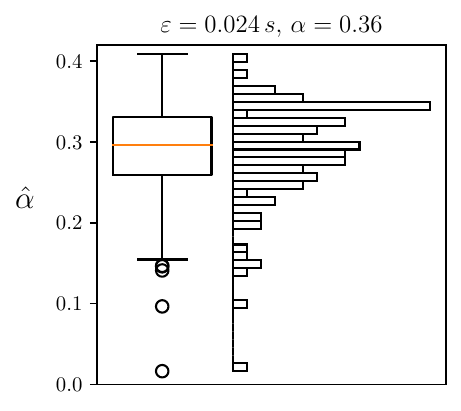}
    \label{fig:long-c}
  \end{subfigure}
\caption{Histograms and box plots of $\hat{\alpha}$ estimates on clips from the BSD dataset with $\varepsilon = \{0.008\,s ,\,0.016\,s,\,0.024\,s\}$. Estimation parameters $\varphi = 5\degree$, $D = 30$. Note that only the (0, 0.4) range is displayed.}
  \label{fig:long_estim}
\end{figure*}

\subsection{Evaluation datasets}
\label{subsec:datasets}
Finding a suitable dataset was difficult, as the exposure time data is often erased from video clip metadata or not saved by the video camera at all. Many popular video datasets such as GoPro \cite{Nah_2017_CVPR} or DeepVideoDeblurring Dataset \cite{su2017deep} are either stripped of this information or are available as individual frames, converted post-capture with camera details unavailable.

The largest public dataset containing exposure time data for every video clip is the Beam-Splitter Dataset (BSD) \cite{zhong2020efficient}. The Beam-Splitter Dataset consists of pairs of identical videoclips captured with different exposure settings by 2 independently controlled cameras. Due to the framerate $f$ being known, we are able to calculate the ground-truth of $\alpha$ for each videoclip. We test on the full 600-videoclip dataset. Exposure parameters for each subset are in \cref{tab:bsd1}. There are 100 videoclips in each distinct subset. We used this dataset for quantitative testing as well as qualitative results on well-performing video clips and failure cases.

\subsection{Estimation of exposure fraction on the BSD dataset}
\label{subsec:quantity_bsd_results}

We performed quantitative evaluation on all subsets of the BSD dataset for all parameter combinations mentioned in \cref{subsec:params}. We use Mean Absolute Error (MAE) as the performance measure: 
\begin{equation}
    \text{\textit{MAE}} = \frac{1}{N}\sum\limits_{i=1}^{N}|\alpha - \hat{\alpha_{i}}|
\end{equation}

Results are presented in \cref{tab:bsd_results_all}. We observe a mild positive relationship between method accuracy (MAE) and the increasing size of selected patches $D$.
Testing also shows that a more relaxed cosine constraint $\varphi$ leads to a lower error
for all fixed patch sizes. We attribute this to the property of the adopted linear blur estimation method, which occasionally produces results with a correct magnitude but incorrect orientation or vice versa, further amplified by its discrete output space \cite{gong2017blur2mf}. 

The estimated $\hat{\alpha}$ is less accurate for very small and very large values of the true exposure fraction $\alpha$. Analysis of the behavior is a part of our future work. We conjecture that for very low values of $\alpha$, the discrete estimates of blur are highly inaccurate. For large values of $\alpha$, the optical flow, operating on blurred images, is possibly losing accuracy.

\cref{fig:short_estim,} and \cref{fig:long_estim} show the distribution of $\hat{\alpha}$ from a test with parameters $\varphi = 5\degree$, $D = 30$. Larger levels of noise are present in the estimates of $\alpha < 0.1$. This supports our conjecture that the accuracy of linear blur kernel estimation on very small values of $\alpha$ is rather low. For values $\alpha > 0.1$, we observe the majority of estimates within close intervals of ground truth values. The dependency of estimation accuracy on the ground truth value of $\alpha$ is the largest limitation of the method. 

\begin{figure*}
  \centering
  \begin{subfigure}{0.45\linewidth}
    \frame{\includegraphics[width=8cm]{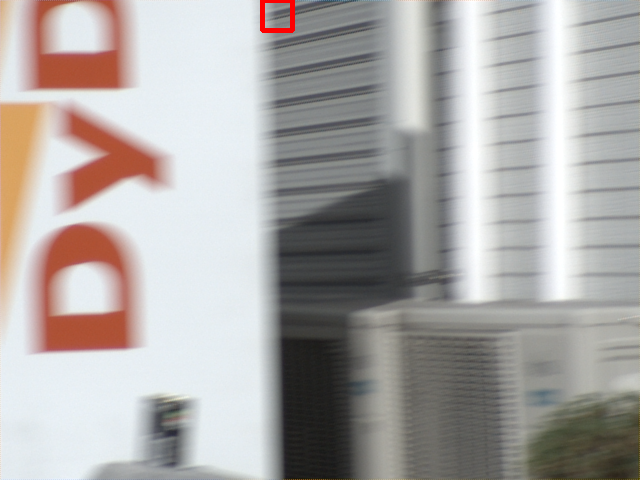}}
    \caption{Video frame $F_{38}$, selected patch highlighted in red.}
    \label{fig:frame_i}
  \end{subfigure}
  \hfill
  \begin{subfigure}{0.45\linewidth}
    \frame{\includegraphics[width=8cm]{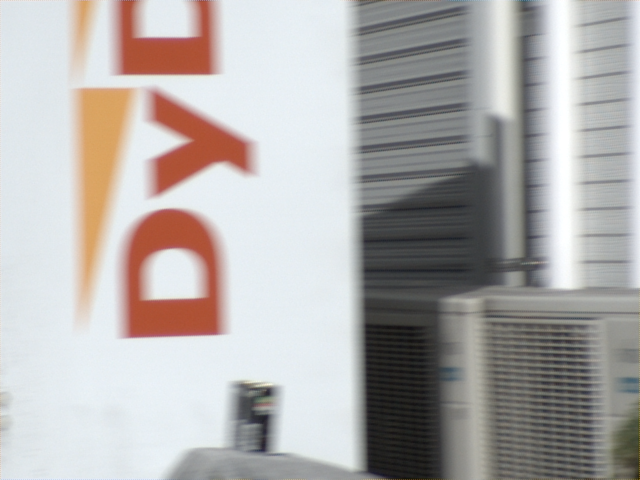}}
    \caption{Video frame $F_{39}$}
    \label{fig:frame_i+1}
  \end{subfigure}
  
    \bigskip
  \begin{subfigure}{0.48\linewidth}
    \centering
    \frame{\includegraphics[width=4cm]{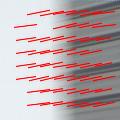}}
    \caption{Linear blur kernels inside and around the selected patch.}
    \label{fig:blur_patch}
  \end{subfigure}
  \hfill
  \begin{subfigure}{0.48\linewidth}
    \centering
    \frame{\includegraphics[width=4cm]{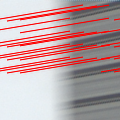}}
    \caption{Optical flow between $F_{38}$, $F_{39}$ inside and around the selected patch.}
    \label{fig:blur_patch}
  \end{subfigure}
\caption{Example patch with nearly perfect agreement with the assumption expressed by \cref{eq:alpha_estimation}. The blur kernel and optical flow estimates are collinear, $\hat{\alpha}_{patch} = 0.26$ and $\alpha = 0.24$. 
The selected patch from frame $F_{38}$,
video clip no. 16 from BSD-16ms subset.
Estimation parameters $\varphi = 5\degree$, $D = 30$. 
}
  \label{fig:ideal}
\end{figure*}

\begin{figure*}
  \centering
  \begin{subfigure}{0.45\linewidth}
    \frame{\includegraphics[width=8cm]{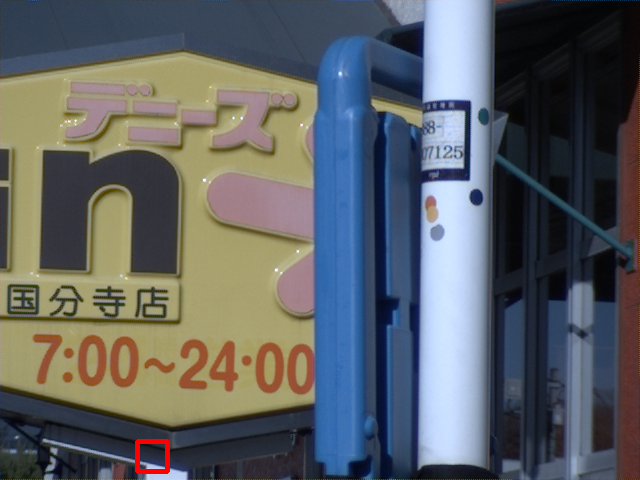}}
    \caption{Video frame $F_{52}$, selected patch highlighted in red.}
    \label{fig:frame_i_failr}
  \end{subfigure}
  \hfill
  \begin{subfigure}{0.45\linewidth}
    \frame{\includegraphics[width=8cm]{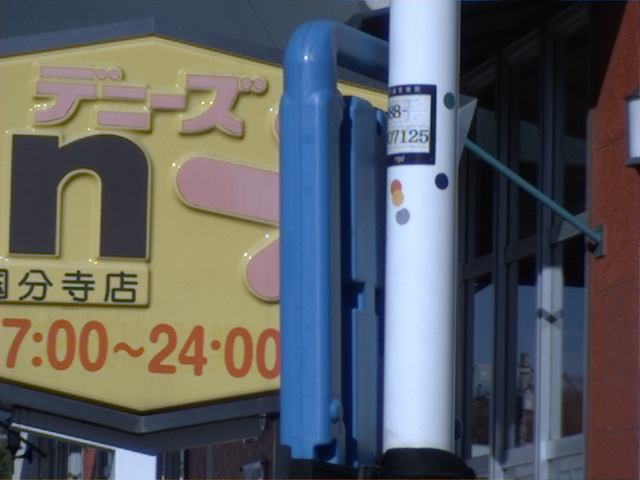}}
    \caption{Video frame $F_{53}$}
    \label{fig:frame_i+1_fail}
  \end{subfigure}
  
    \bigskip
  \begin{subfigure}{0.48\linewidth}
    \centering
    \frame{\includegraphics[width=4cm]{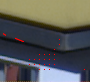}}
    \caption{Linear blur kernels inside and around the selected patch.}
    \label{fig:blur_patch_fail}
  \end{subfigure}
  \hfill
  \begin{subfigure}{0.48\linewidth}
    \centering
    \frame{\includegraphics[width=4cm]{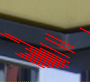}}
    \caption{Optical flow between $F_{52}$, $F_{53}$ inside and around the selected patch.}
    \label{fig:flow_patch_Fail}
  \end{subfigure}
\caption{A failure case of linear blur kernel estimates in a dark, low contrast area with no texture; $\hat{\alpha}_{patch} = 0.065$, $\alpha = 0.015$. The linear blur kernel estimator fails to accurately model the blur magnitudes, resulting in an inaccurate estimate of $\hat{\alpha}_{patch}$. Since the kernels still satisfy the orientation and magnitude constraints defined in \cref{eq:cosine_constraint}, \cref{eq:blur_flow_length_constraints}, \cref{eq:minimum_length_constraint}, they are considered valid. Similar situations remain challenging for both the linear blur kernel estimator and the method. Frame $F_{52}$, estimation parameters $\varphi = 5\degree$, $D = 30$. Video clip no. 74 from BSD-1ms subset.}
  \label{fig:fail}
\end{figure*}

\subsection{Qualitative analysis of results on BSD dataset}
\label{subsec:quality_bsd_results}

From the estimates performed with parameters $\varphi = 5\degree$, $D = 30$ detailed in \cref{fig:short_estim}, \cref{fig:long_estim}, we selected two videoclips for in-depth analysis in an attempt to further compare the values of linear blur kernel estimates and optical flow, and their effect on per-frame estimates of $\alpha$.

As an example of the ideal case, we selected videoclip no. 16 from the BSD-16ms subset. The estimated value $\hat{\alpha} = 0.22$; ground truth $\alpha = 0.24$.
In \cref{fig:ideal}, we display frames $F_{38}, \,F_{39}$, the patch selected by the method, and detailed visualization of both linear blur kernel estimates and optical flow vectors. In this ideal case, we observe near-perfect collinearity of linear blur kernels and optical flow vectors, as well as relatively uniform magnitudes of both vectors. We attribute the good performance of both linear blur kernel estimation and optical flow to the largely linear, lateral motion of the camera and the presence of areas with blurred textured surfaces in the scene. Similar situations with camera motion are ideal for the utilized linear blur kernel estimator, as Gong \etal's method \cite{gong2017blur2mf} was trained on synthetic data modeled as blur by camera motion.

We also analyze an example of a failure case in \cref{fig:fail} where the estimate $\hat{\alpha}$ is grossly erroneous (ground truth $\alpha = 0.015$, estimate $\hat{\alpha} = 0.065$). Here, we observe an incorrectly estimated magnitude of linear blur kernels, resulting in an overestimate of $\alpha$. The method of \cite{gong2017blur2mf} seems to fail in dark areas with low contrast and no pronounced textures.
As discussed in \cref{subsec:quantity_bsd_results}, the linear blur kernel estimator does not estimate low levels of motion blur accurately.
In consequence, the method often fails to produce accurate estimates for ground truth values of $\alpha < 0.1$.

\begin{table}
\renewcommand{\arraystretch}{1.2}
  \centering
  \begin{tabular}{ccccccc}
    \toprule
    Clip \# & $k$ & $\alpha$ & $\hat{\alpha}$ & $\alpha^\prime$ & $\hat{\alpha}^\prime$ & Abs. error  \\
    \midrule
    78  & $3$ & $0.360$ & $ 0.349$ & $0.120$ & $0.117$ & $0.003$\\
    14  & $3$ & $0.240$ & $0.230$ & $0.080$ & $0.075$ & $0.005$\\
    115 & $2$ & $0.120$ & $0.121$ & $0.060$ & $0.054$ & $0.006$\\
    \bottomrule
  \end{tabular}
  \caption{Detection of video clip subsampling by integer factor~$k$. The ground truth $\alpha$
  and estimated $\hat{\alpha}$ on the original sequence and 
   the GT $\alpha^\prime$ and the estimated $\hat{\alpha}^\prime$ exposure fractions on the tampered video. In all cases, the value of $\hat{\alpha}^\prime$ was estimated accurately. The test was performed on videoclips 78 (BSD-24ms), 14 (BSD-16ms), and 115 (BSD-8ms) containing traffic and moving vehicles. Estimation parameters $\varphi = 5\degree$, $D = 30$.}
  \label{tab:frame_deletion}
\end{table}

\subsection{Detection of video alteration by frame deletion}
\label{subsec:frame_deletion}
Video frame deletion is a form of video clip tampering that directly affects the temporal consistency introduced by the camera physical properties and its exposure mechanism. When consecutive video frames are deleted, objects in the scene exhibit progressively larger inter-frame motions, proportionally to the number of removed frames, yet the motion blur remains the same. If the time scale, i.e. the playback frame rate, is edited or ignored by the player, the replayed video will appear to the viewer to have faster motions. As a result of frame deletion, the estimated value of $\alpha$ will not be consistent with the original video clip; the tampered section will have values of the $\alpha$ different from the rest of the video.

Frame deletion and insertion may be used for malicious purposes in video clips where the speed of motion provides significant information value, such as video clips from automotive dash cameras that could be used for speed measurements. Similarly, it may be utilized to remove frames containing sensitive or identifying information, such as license plates or faces on footage from surveillance cameras. In the case of dash and surveillance cameras, the capture framerate $f$ and $\varepsilon$ are often available as camera metadata, allowing a comparison between altered video clips and ground truth~$\alpha$.

We selected three video clips of scenes containing traffic and moving vehicles with  $\alpha > 0.1$. This choice was based on the fact that the method performs better on larger values of $\alpha$ due to the limitations of the linear blur estimator.
For each of the selected video clips, we performed frame subsampling with an integer factor $k$, i.e. every $k\text{-th}$ frame was preserved; the intermediate frames were discarded. The new apparent ground truth value of $\alpha^\prime = \frac{\alpha}{k}$ where $\alpha$ is the ground truth value of the source video clip. The results are displayed in \cref{tab:frame_deletion}. For all subsampled videoclips, the method produced estimates within a close margin of the apparent ground truth value. Note significantly lower error than on video clips where unmodified $\alpha = \{0.015,\,0.030,\,0.045\}$. We attribute this to more accurate estimates in the selected videoclips, as their unmodified $\alpha = \{0.120,\,0.240,\,0.360\}$ results in larger motion blur and therefore more accurate linear blur kernel estimates.
\begin{table}
\renewcommand{\arraystretch}{1.2}
  \centering
  \begin{tabular}{ccccc}
    \toprule
    Clip \# & Interpolation factor & $\alpha$ & $\hat{\alpha}$  & $\hat{\alpha}^\prime$  \\
    \midrule
    78  & 2x & $0.360$ & $ 0.349$ &  $0.668$ \\
      & 4x & -- & -- & $0.536$ \\
    14  & 2x & $0.240$ & $0.230$ & $0.412$ \\
      & 4x & -- &  -- & $0.562$\\
    \bottomrule
  \end{tabular}
  \caption{Detection of video clip interpolation. The ground truth $\alpha$ and estimated $\hat{\alpha}$ on the original sequence and the estimated $\hat{\alpha}^\prime$ exposure fractions on the tampered video. In all cases, the value of $\hat{\alpha}^\prime$ increased noticeably. The test was performed on videoclips 78 (BSD-24ms) and 14 (BSD-16ms) containing traffic and moving vehicles. Estimation parameters $\varphi = 5\degree$, $D = 30$}
  \label{tab:frame_interpolation}
\end{table}
\subsection{Detection of video alteration by frame interpolation}
\label{subsec:frame_interpolation}
Video frame interpolation is a technique of temporal alteration that synthesizes new intermediate frames in order to increase video framerate and for motion to appear smoother. It may be followed by a change of playback timescale, in which case it results in a slow-motion video. A videoclip altered in such a way can then be used as fake evidence of vehicle speed from a surveillance or dash camera. Techniques derived from video frame interpolation may also be used to add new content to videos or change their appearance, such as interpolation between still photographs of a person's face for the creation of so-called "deepfakes" \cite{8639163}. As described in \cref{subsec:frame_deletion}, surveillance or dash cameras often save the value of $\varepsilon$ and $f$ by directly imprinting it on video frames or by saving it to the metadata. This provides the ground-truth value of $\alpha$ for comparison with the method estimate.

 Based on the definition of exposure fraction $\alpha$ (\cref{subsec:shutter_formulation}), it is expected its value to increase if the newly-synthesized intermediate frames reduce the inter-frame motion of objects without affecting the amount of motion blur, i.e. the newly synthesized frames appear as blurry as the source frames.  It is, however, not possible to compute the value of $\alpha^\prime$ of interpolated videoclips accurately, as modern deep neural network-based interpolation methods such as RIFE \cite{huang2022rife} do not perform parametrized blurring or deblurring.

We performed interpolation on videoclips no. 78 and 14 from the experiment in \cref{subsec:frame_deletion} with the state-of-the-art RIFE method \cite{huang2022rife}.
For each videoclip, we tested 2x interpolation and 4x interpolation. Results are presented in \cref{tab:frame_interpolation}. We observe an increase in estimates of $\alpha^\prime$ in all cases of interpolation, pointing to the method synthesizing new frames with similar amounts of motion blur as the source. 
 
\section{Further applications}
\label{sec:other_appls}
In applications concerning video frame interpolation and video frame deblurring, the value of $\alpha$ might help parameterize blur for more accurate deblurring or motion modeling. The synthetization of new frames from a blurry source remains a challenge for modern interpolation methods, and accurate blur modeling might provide the necessary information for performance improvements \cite{Ji_2023_ICCV}. In the case of linear blur estimates and optical flow, the estimate of $\alpha$ is useful as a complement in computing values for positions $(x, \,y)$ where one of the methods failed (under the assumption that is it possible to estimate the value of $\alpha$ from other frames and positions in the videoclip). This might be useful for the creation of new datasets for linear blur kernel estimation or optical flow, as the parameters may be estimated from existing datasets and computed for entire frames or videoclips.

\section{Conclusion}
\label{sec:conclusion}
We proposed a novel method for estimating the exposure fraction based on dense optical flow and linear blur estimates.
The method was evaluated on the publicly available BSD Dataset. The mean absolute error was $0.039$; the method performed best in the range $(0.12,\,0.36)$. We observed reduced accuracy for ground truth values below $0.1$, leading us to conjecture that the use of discrete linear blur kernel estimates may be a limiting factor. Developing an improved method for the estimation of linear blur kernels is a key part of our future work.
Lastly, we presented a possible application of exposure fraction estimation for video tampering detection, specifically of frame deletion and frame insertion. 
The implementation is available at \url{https://github.com/edavidk7/exposure_fraction_estimation}.

\noindent
 {\bf Acknowledgement.} The authors were supported by the Research Center for Informatics project CZ.02.1.01/0.0/0.0/16\_019/0000765 of OP VVV MEYS.
\vspace{-3ex} 
{\small
\bibliographystyle{ieee_fullname}

}

\end{document}